# Unsupervised Learning of Monocular Depth and Ego-Motion Using Multiple Masks

Guangming Wang, Hesheng Wang, Yiling Liu and Weidong Chen

*Abstract*— A new unsupervised learning method of depth and ego-motion using multiple masks from monocular video is proposed in this paper. The depth estimation network and the ego-motion estimation network are trained according to the constraints of depth and ego-motion without truth values. The main contribution of our method is to carefully consider the occlusion of the pixels generated when the adjacent frames are projected to each other, and the blank problem generated in the projection target imaging plane. Two fine masks are designed to solve most of the image pixel mismatch caused by the movement of the camera. In addition, some relatively rare circumstances are considered, and repeated masking is proposed. To some extent, the method is to use a geometric relationship to filter the mismatched pixels for training, making unsupervised learning more efficient and accurate. The experiments on KITTI dataset show our method achieves good performance in terms of depth and ego-motion. The generalization capability of our method is demonstrated by training on the low-quality uncalibrated bike video dataset and evaluating on KITTI dataset, and the results are still good.

## I. INTRODUCTION

In the fields of mobile robots and autonomous driving, Simultaneous Localization and Mapping (SLAM) is a basic key technology, and its main task is to build accurate environmental maps around the robot and estimate its ego-motion. This is crucial for the next trajectory planning and decision making. However, the model-based method faces many difficulties. The monocular needs to be initialized and the scale needs to be calibrated. Feature-Based visual SLAM faces problems such as loss of matching features in dynamic environments and few features sometimes.

Learning-based method has the potentials to handle these problems for its robustness to a variety of challenging environments. Tateno et al. [1] used deep learning to predict the depth of key frames, and adjacent frames were used for depth detail adjustment. The method achieves a good monocular performance without initialization and shows the superiority of learning-based depth estimation in SLAM. The supervised method of depth estimation Tateno K et al. used comes from [2]. The supervised learning was also adopted by [3] and [4] for depth estimation. However, the acquisition of most data requires a lot of equipment for observation and calibration settings, which severely limits the training data that can be used by supervised methods. Moreover, the calibration and acquisition of these data are not completely accurate inherently such as manual labeling. The methods of unsupervised learning can get rid of these problems.

In this paper, a new unsupervised learning method using multiple masks for monocular depth and ego-motion estimation is proposed. Its innovation and effectiveness are as follows:

- The process of image reconstruction is carefully analyzed, and according to the pixel mismatch problem in the calculation of the loss function, the overlap mask and blank mask on the projected image and the target image are designed corresponding to the "overlap" and "blank" conditions to solve it.

- After three masks, some details are further considered, and repeated masking is proposed. A more detailed mask is made, which naturally solves the problem of image pixel mismatch.

- The mismatched pixels are carefully masked and filtered for the calculation of loss functions by geometric relationship, which makes the backpropagation in the unsupervised learning process more accurate. The superiority of the method is demonstrated on KITTI dataset [5], Cityscapes dataset [6] and Uncalibrated Bike Video Dataset [7].

The rest of this article is organized as follows. The Section II summarizes some works related to this paper. Section III describes masks for image reconstruction. Section IV introduces the architecture and loss function of our system. Section V shows the experiment results and the effect of the masks proposed by us. Finally, the conclusion is drawn in the sixth section.

## II. RELATED WORK

Unsupervised learning of depth begins with binoculars for training. Garg et al. [8] used a binocular image pair without the ground truth of depth, where depth is generated as an intermediate output and the supervision of the network is from the other image. Godard et al. [9] additionally considered depth smoothness loss and left-right consistency. Binocular sequences with deep feature-based warping loss were used in [10] for learning depth and odometry. The use of binocular sequences enables the use of spatial and temporal photometric errors and limits scene depth and camera motion to real world scales. Li et al. [11] proposed a monocular visual odometry system called UnDeepVO that recovers the scale by training the UnDeepVO using a binocular image pair but tests it by using continuous monocular images.

For the first time, Zhou et al. [12] conducted an unsupervised joint estimation of depth and ego-motion from monocular video and obtained competitive depth and pose

*This work was supported in part by the Natural Science Foundation of China under Grant U1613218 and 61722309. Corresponding Author: Hesheng Wang.

G. Wang, H. Wang, Y. Liu and W. Chen are with Department of Automation, Shanghai Jiao Tong University, Shanghai 200240, China and Key Laboratory of System Control and Information Processing, Ministry of Education of China.

results. This further reduces the cost of data acquisition, resulting in a significant increase in the amount of data that can be used for learning. Mahjourian et al. [7] then added spatial geometric information of Iterative Closest Point (ICP) [13]-[15] to improve the results. Yin et al. [16] deal with static and dynamic scene parts separately and jointly estimate depth, optical flow and camera pose. Yang et al. [17] introduces a 3D-as-smooth-as-possible (3D-ASAP) as prior information in the method, which can jointly estimate the edge, normal and depth for fine detailed structures. Wang et al. [18] trained the depth network using direct visual odometry rather than pose network. They also found a depth normalization strategy to improve the performance of depth estimation. Unsupervised learning is making an increasing contribution to SLAM.

## III. THE MASKS FOR IMAGE RECONSTRUCTION

Our approach learns depth and ego-motion from monocular video without ground-truth. Fig. 1 illustrates the main part of our contribution, where there are two new masks, $M_{ove}$ and $M_{bla}$. The geometric relationship of image reconstruction between adjacent frames is carefully considered, and two masks are designed to naturally screen out the wrong matching pixels. Note that the depth and ego-motion are estimated by the depth estimation network and the pose estimation network, two independent networks for depth estimation of single frame and ego-motion estimation of adjacent frames. The two losses, image reconstruction loss and structural similarity (SSIM) loss are used to measure the difference between the reconstructed image and the original image for the unsupervised training of the networks. This section first describes the process of image reconstruction, and then naturally introduces two important masks proposed by us. The depth estimation network and pose estimation network as well as two loss function will be further clarified in IV section.

### A. Image Reconstruction

The image reconstruction process can be described as follows: giving a pair of consecutive frame images $X_{t-1}$ and $X_t$. The image depth $D_t$ at the time $t$ and the ego-motion $T_t$ of the camera from time $t-1$ to $t$ have been estimated by the depth network and the pose network. Then, $\hat{X}_t$ can be reconstructed from the pixel values of $X_{t-1}$.

First, the pixel $(i, j)$ with estimated depth $D_t^{ij}$ in image $X_t$ can be projected as a 3D point cloud:

$$Q_t^{ij} = D_t^{ij} \cdot K^{-1}[i, j, 1]^T \quad (1)$$

Where $K$ is the camera intrinsic matrix, with homogeneous coordinates.

Next, $Q_t$ can be converted to a point cloud for the previous frame with the estimated motion $T_t$ of the camera from $t-1$ to $t$:

$$\hat{Q}_{t-1} = T_t Q_t \quad (2)$$

Then the point cloud $\hat{Q}_{t-1}$ can be projected onto the imaging plane at frame $t-1$ to get $K\hat{Q}_{t-1}$. The corresponding point after projection is recorded as $(\hat{i}, \hat{j})$. The above transformation relationship can be written as:

$$[\hat{i}, \hat{j}, 1]^T = KT_t(D_t^{ij} \cdot K^{-1}[i, j, 1]^T) \quad (3)$$

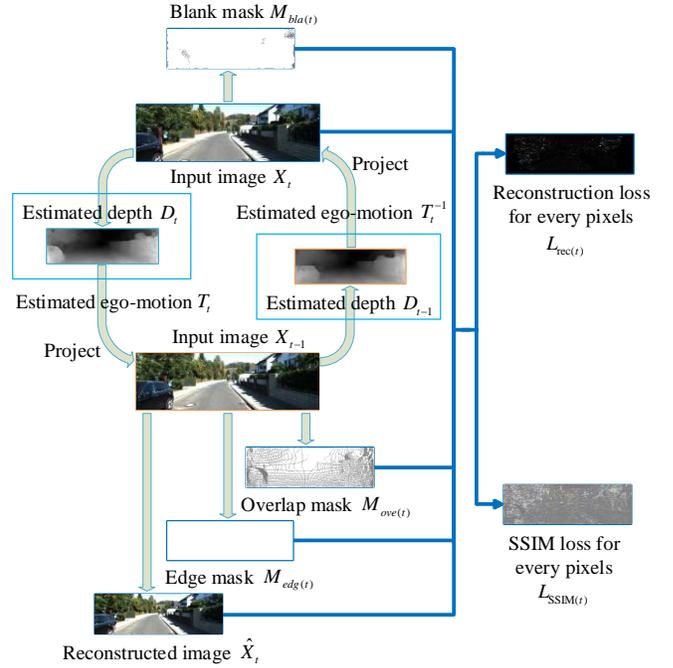

Fig. 1: Computational relationship of the main loss functions for network training. Our system uses consecutive two frames from monocular video to reconstruct the images based on the depth and ego-motion estimated by the depth estimation network and the pose estimation network, obtaining several masks at the same time. Next, the reconstructed image, the original image, and several masks are used to calculate the two main loss functions for backpropagation. The Figure just shows the process in one direction (*t*) and the other direction (*t*-1) is also used and similar.

In order to get the reconstructed $\hat{X}_t$, it is necessary to obtain each pixel value of the reconstructed $t$ frame according to the pixel value of corresponding position projected to $t-1$ frame. Since the coordinate value projected to the $t-1$ frame is not an integer, the differentiable bilinear sampling mechanism [19] at the projected position is taken next.

As shown in Fig. 2, coordinate $\hat{p}_t$ on the imaging plane of $X_{t-1}$ can be projected from the coordinate $p_t$ of the pixel on $X_t$. In the bilinear interpolation method, the corresponding pixel value $\hat{X}_t(p_t)$ on the reconstructed image $\hat{X}_t$ has the following representation:

$$\hat{X}_t(p_t) = \sum_{i \in \{t,b\}, j \in \{l,r\}} w^{ij} X_{t-1}(p_{t-1}^{ij}) \quad (4)$$

Where $w^{ij}$ is the proportional term for bilinear interpolation, measuring the spatial proximity of $\hat{p}_t$ and $p_{t-1}^{ij}$, with $\sum_{i,j} w^{ij} = 1$. In the end, the reconstructed image $\hat{X}_t$ is gotten. This process is also repeated in the other direction. And we can get the reconstructed image $\hat{X}_{t-1}$.

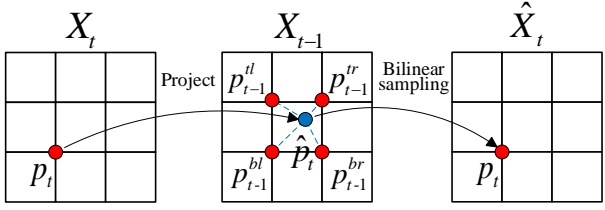

Fig. 2: The bilinear interpolation mechanism used in image reconstruction. Each pixel in the reconstructed image is usually calculated from four pixels of the adjacent frame.

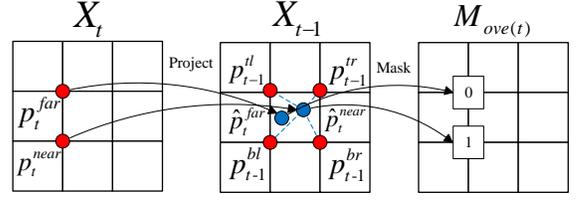

Fig. 4: The mechanism of overlap mask in the image reconstruction process.

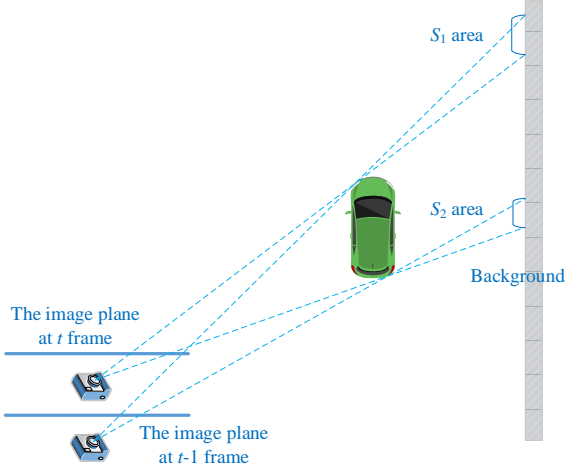

Fig. 3: The problem of pixel mismatch in image reconstruction due to camera motion. This figure aids in clarifying the reasons why the two masks are naturally considered. The closer the obstacle is to the camera, the greater the impact will be.

### B. The Overlap Mask on the Projected Image

The image $X_{t-1}$ becomes image $X_t$ as the camera moves from time $t-1$ to $t$. Some of the pixels originally in image $X_{t-1}$ leave in $X_t$, and some pixels not originally in image $X_{t-1}$ enter image $X_t$. Take the case of projecting $t$ frame coordinates to image plane at $t-1$ frame as an example. The difference is not limited to the principled edge mask in [7] and [17], which means that the pixel position at $t$ frame is masked when the pixel coordinate in the $t$ frame is projected outside image plane at $t-1$ frame. The principled edge mask is symbolized as $M_{edg}$ in this paper.

As shown in Fig. 3, for convenience of observation and explanation, the camera motion from $t-1$ to $t$ frame is presented in a two-dimensional plane. In the image $X_{t-1}$ at $t-1$ frame, the background of the $S_1$ area cannot be seen due to the occlusion of the car, but can be seen at $t$ frame. Similarly, the $S_2$ area that is not visible in $t$ frame can be seen at $t-1$ frames. This causes an overlap problem when pixel coordinates are projected onto each other. The pixels at the front part of the car and the $S_1$ area will overlap when the pixel coordinates on $X_t$ are projected to the image plane at $t-1$ frame. At this point, it needs to be decided which pixels to match with the pixels in $X_{t-1}$. As we can see in Fig. 3, the camera can only see the front part of the car at $t-1$ frame and can't see the area $S_1$. Therefore, when pixel overlap occurs, the pixel whose 3D point is closer to the imaging plane should be selected to participate in pixel matching, while the far pixel should be masked.

The overlap problem at the pixel level is shown in Fig. 4. We define the overlap occurs when the pixel coordinates are projected into a "grid" that requires four identical interpolation points. The far pixel should be masked, that is, the corresponding pixel on $X_t$ and $\hat{X}_t$ is multiplied by 0 and does not participate in the calculation of loss functions. The overlap mask is denoted as $M_{ove}$, which is a matrix where the position at masked pixel is 0 and a non-masked position is 1.

### C. The Blank Mask on the Target Imaging Plane to Project

As shown in Fig. 3, a blank area corresponding to $S_2$ area will be generated on the $X_{t-1}$ when the pixel coordinates at $t$ frame are projected to the imaging plane at $t-1$ frame. This blank area is not used when calculating the reconstruction loss for the reconstructed image $\hat{X}_t$ and $X_t$. But it will be used when the coordinates at $t-1$ frame are projected onto the imaging plane at $t$ frame to calculate the reconstruction loss of the $t-1$ frame. In fact, this part will be masked by overlap mask when $t-1$ frame is projected to $t$ frame, but this is not always repeated. In the subsequent experiments it can also be seen that the effects of the two masks are quite different due to the discrete nature of pixels.

As shown in Fig. 5, the camera can see the $S_1$ area at $t-1$ frame, but $S_1$ area will be blocked by the car at $t$ frame. Therefore, when the pixel coordinates on $X_{t-1}$ are projected to the imaging plane at $t$ frame to reconstruct $\hat{X}_{t-1}$, the pixels corresponding to $S_1$ region on $X_{t-1}$ will be calculated with the pixels of the right side of the car on $\hat{X}_{t-1}$ reconstructed from $X_t$. This is an incorrect match, and this part of computational loss is a kind of interference to the training of the neural network. Moreover, the situation cannot be eliminated by the projection from $t-1$ to $t$ frame. But if back projection is considered, this part of erroneous correspondence can be eliminated. When the projection is from $t$ to $t-1$ frame, a blank is generated in the area corresponding to the $S_1$ area on $X_{t-1}$. So, the image reconstruction loss for the reconstructed image $\hat{X}_{t-1}$ and $X_{t-1}$ should be calculated after the pixels corresponding $S_1$ area on $X_{t-1}$ is masked by reverse projection. Therefore, the two-way reconstruction loss is calculated after all the masks have been calculated.

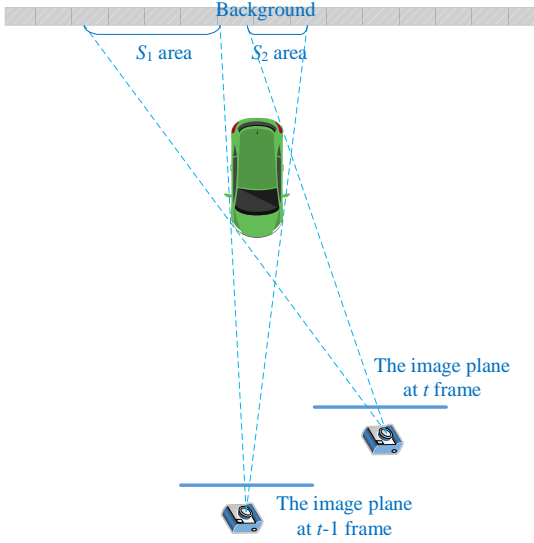

Fig. 5: One condition of pixel mismatch in image reconstruction for $t-1$ frame that cannot be solved by overlap mask.

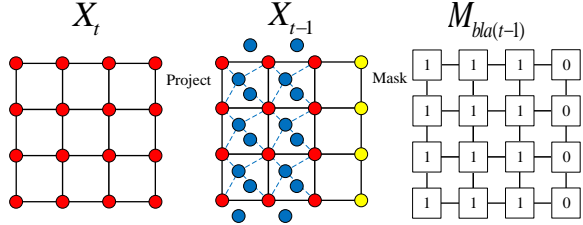

Fig. 6: The mechanism of blank mask in the image reconstruction process.

The blank mask is shown in Fig. 6 at the pixel level. When the pixel coordinates at $t$ frame is projected to the $t-1$ frame, we define the coordinate position at the $t-1$ frame that does not participate in any interpolation contribution as the "blank area", that is, the position corresponding to "blank area" on $M_{bla}$ is set to 0, and the other position is 1, forming the blank mask $M_{bla}$.

### D. One Special Case towards Repeating Mask

The mismatch problem in most cases can be solved by the three masks above, but still not for all cases. When there is a thin object in the scene, or the camera moves fast, the situation shown in Fig. 7 may occur. This situation is relatively rare, and perhaps this section will be more useful when depth and pose estimates are more accurate in the future.

In Fig. 7, the camera moves so fast that two different sides of the car are seen in two consecutive frames. The two-way projection is performed to generate the three masks, the edge mask, the overlap mask, and the blank mask. After the masks, there will still be a mismatched part. When the pixel coordinates at $t-1$ frame are projected to the $t$ frame, the coordinates corresponding to $S_1$ area on the left side of the car will be projected to the area corresponding to $S_2$ area on the right side of the car in the imaging plane at $t$ frame. And this kind of error can't be solved directly by any of the three masks.

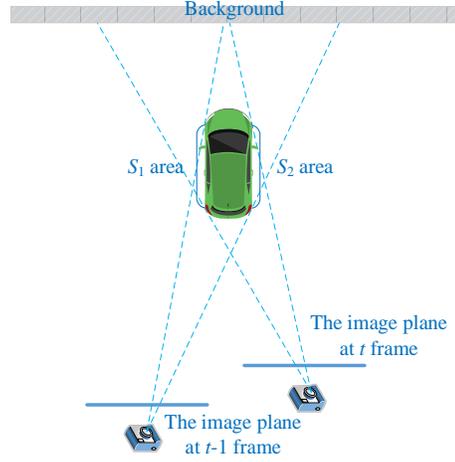

Fig. 7: One condition of pixel mismatch that cannot be directly solved by the three masks. The object may not be a car, but a thin object such as a billboard standing on the ground.

However, due to the overlap mask and blank mask, most mismatched pixels on the background has been eliminated. At this point, projecting the $t-1$ frame to the imaging plane at $t$ frame will reveal a new "blank area" in the imaging plane at $t$ frame, which can mask new mismatched pixels with the blank mask. And the mutual projection and masking operations are performed back and forth until the "blank areas" is no longer present, and the pixels at the two frames can completely correspond exactly. The thinner the object is, the less effect the two-way projection makes for each time. Three times of the two-way projection are taken in the experiments.

## IV. TRAINING SETTINGS FOR UNSUPERVISED TRAINING

Three loss functions are adopted in the system. Due to the introduction of these masks, the masked pixels are deleted from the loss function calculation.

### A. Image Reconstruction Loss

Comparing the reconstructed image with the corresponding real image will produce a differentiable image reconstruction loss function based on photometric consistency [8], [12], which is formulated as[1]:

$$L_{rec} = \sum_{ij} \left\| (X_t^{ij} - \hat{X}_t^{ij}) M_{edg(t)}^{ij} M_{ove(t)}^{ij} M_{bla(t)}^{ij} \right\| \quad (5)$$

### B. Similarity Loss and Smoothness Loss

Similar to [9] and [20], structural similarity (SSIM) is adopted in our method as a term of loss function in the training process, which is a common measure for evaluating the quality of image prediction. And the loss function is:

$$L_{SSIM} = \sum_{ij} [1 - SSIM(X_t^{ij}, \hat{X}_t^{ij})] M_{edg(t)}^{ij} M_{ove(t)}^{ij} M_{bla(t)}^{ij} \quad (6)$$

---

[1] Both image reconstruction loss and SSIM loss include two-way image reconstruction for calculations. But only $t$ frame is reserve in the formula for brevity.

TABLE I: Monocular depth evaluation results tested on KITTI Eigen et al. [3] test split. Since Garg et al. [8] only provide the depth results cap of 50m, of which the comparison is also performed in our experiments. Other results all have an upper limit of 80m. CS + K means fine tuning on KITTI dataset [5] after pre-training on Cityscapes dataset [6]. Bike means training only on Uncalibrated Bike Video Dataset [7]. K means training only on KITTI dataset [5]. DN means depth normalization from [18]. The best and the second best performance in each block are highlighted in BLUE and GREEN.

| Method | Supervised | Dataset | Error metric | | | | Accuracy metric | | |
|---|---|---|---|---|---|---|---|---|---|
| | | | Abs Rel | Sq Rel | RMSE | RMSE log | $\delta < 1.25$ | $\delta < 1.25^2$ | $\delta < 1.25^3$ |
| Eigen et al. [3] Coarse | Depth | K | 0.214 | 1.605 | 6.563 | 0.292 | 0.673 | 0.884 | 0.957 |
| Eigen et al. [3] Fine | Depth | K | 0.203 | 1.548 | 6.307 | 0.282 | 0.702 | 0.890 | 0.958 |
| Liu et al. [4] | Depth | K | 0.202 | 1.614 | 6.523 | 0.275 | 0.678 | 0.895 | 0.965 |
| Godard et al. [9] | Stereo | K | 0.148 | 1.344 | 5.927 | 0.247 | 0.803 | 0.922 | 0.964 |
| Zhan et al. [10] | Stereo | K | 0.144 | 1.391 | 5.869 | 0.241 | 0.803 | 0.928 | 0.969 |
| Zhou et al. [12] | No | K | 0.208 | 1.768 | 6.856 | 0.283 | 0.678 | 0.885 | 0.957 |
| GeoNet [16] | No | K | 0.164 | 1.303 | 6.090 | 0.247 | 0.765 | 0.919 | 0.968 |
| Mahjourian et al. [7] | No | K | 0.163 | 1.240 | 6.220 | 0.250 | 0.762 | 0.916 | 0.968 |
| LEGO [17] | No | K | 0.162 | 1.352 | 6.276 | 0.252 | - | - | - |
| Ours | No | K | 0.158 | 1.277 | 5.858 | 0.233 | 0.785 | 0.929 | 0.973 |
| Ours +DN | No | K | 0.154 | 1.163 | 5.700 | 0.229 | 0.792 | 0.932 | 0.974 |
| Garg et al. [8] cap 50m | Stereo | K | 0.169 | 1.080 | 5.104 | 0.273 | 0.740 | 0.904 | 0.962 |
| Godard et al. [9] cap 50m | Stereo | K | 0.140 | 0.976 | 4.471 | 0.232 | 0.818 | 0.931 | 0.969 |
| Zhan et al. [10] cap 50m | Stereo | K | 0.135 | 0.905 | 4.366 | 0.225 | 0.818 | 0.937 | 0.973 |
| Zhou et al. [12] cap 50m | No | K | 0.201 | 1.391 | 5.181 | 0.264 | 0.696 | 0.900 | 0.966 |
| GeoNet [16] cap 50m | No | K | 0.157 | 0.990 | 4.600 | 0.231 | 0.781 | 0.931 | 0.974 |
| Mahjourian et al. [7] cap 50m | No | K | 0.155 | 0.927 | 4.549 | 0.231 | 0.781 | 0.931 | 0.975 |
| Ours cap 50m | No | K | 0.150 | 0.958 | 4.406 | 0.218 | 0.801 | 0.939 | 0.978 |
| Ours +DN cap 50m | No | K | 0.147 | 0.889 | 4.290 | 0.214 | 0.808 | 0.942 | 0.979 |
| Godard et al. [9] | Stereo | CS + K | 0.124 | 1.076 | 5.311 | 0.219 | 0.847 | 0.942 | 0.973 |
| Zhou et al. [12] | No | CS + K | 0.198 | 1.836 | 6.565 | 0.275 | 0.718 | 0.901 | 0.960 |
| LEGO [17] | No | CS + K | 0.159 | 1.345 | 6.254 | 0.247 | - | - | - |
| Mahjourian et al. [7] | No | CS + K | 0.159 | 1.231 | 5.912 | 0.243 | 0.784 | 0.923 | 0.970 |
| Ours | No | CS + K | 0.155 | 1.184 | 5.765 | 0.229 | 0.790 | 0.933 | 0.975 |
| Mahjourian et al. [7] | No | Bike | 0.211 | 1.771 | 7.741 | 0.309 | 0.652 | 0.862 | 0.942 |
| Ours | No | Bike | 0.204 | 1.645 | 6.939 | 0.293 | 0.676 | 0.878 | 0.950 |

Where SSIM denotes the structural similarity index for two images in [21]. We also use the common smoothness loss as in [7] and [16], symbolized as $L_{smooth}$ in this paper.

C. Training Settings

All loss functions are applied to four different scales $l$. The complete loss is:

$$L = \sum_l \alpha L_{rec}^l + \beta L_{smooth}^l + \gamma L_{SSIM}^l \quad (7)$$

Where $\alpha, \beta, \gamma$ are hyperparameters, which we set to $\alpha=0.15, \beta=0.03, \gamma=0.85$. And we change $\beta=0.2$ for smoothness loss when adding the depth normalization [18].

SfMLearner architecture [12] is adopted as backbone of the network. The architecture consists of two networks. A depth estimation network receives a single image as input and produces a dense depth estimate. Another pose estimation network, which receives successive three frames of images, produces an estimate of camera ego-motion for each successive two frames.

During our unsupervised training, three consecutive frames are as inputs. Then, the pose of consecutive frames and the depth of each frame are predicted by the depth network and pose network. Next, the six masks and two reconstructed images are gotten. Afterwards, the loss functions are calculated from the six masks, two reconstructed images and two source images. Finally, the network parameters are continuously adjusted by backpropagation of the loss functions and a training cycle is completed.

The Adam optimizer [22] is adopted in our training process, $\beta_1 = 0.9, \beta_2 = 0.999$, the learning rate $\alpha=0.0002$. Our depth network uses batch normalization [23], which is not used by the pose network. With small batch training, each small batch is set to 4.

V. EXPERIMENT AND EVALUATION

Experiments are conducted mainly on the KITTI dataset [5], which is the most common data set for evaluating depth and ego-motion. Since [12], [7], [17] all report the results pre-trained on the Cityscapes dataset [6], this part is also added in our results for comparison. Our model is also trained on the uncalibrated bike video dataset [7] and tested on KITTI to show the generalization ability. The experiment results are compared with [7].

A. Evaluation of Depth Estimation

Table I quantitatively compares the depth estimation results of ours and other monocular methods, where lower numbers are better for error metrics and higher numbers are better for accuracy metrics. In the table, [9], [11] and [10] are all trained with binocular information, which is equivalent to providing pose supervision between images. Compared with [3] and [4], our method does not have the supervision for depth. Compared with [7], our method dose not adopt complex ICP back propagation. Compared with [16], our method does not use edge network as an aid. As shown in Table I, our method has surpassed the unsupervised methods and is also competitive with supervised methods. In addition, the results of our method with the depth normalization [18] are also reported in Table I. The depth normalization improves the performance and make the train process more stable.

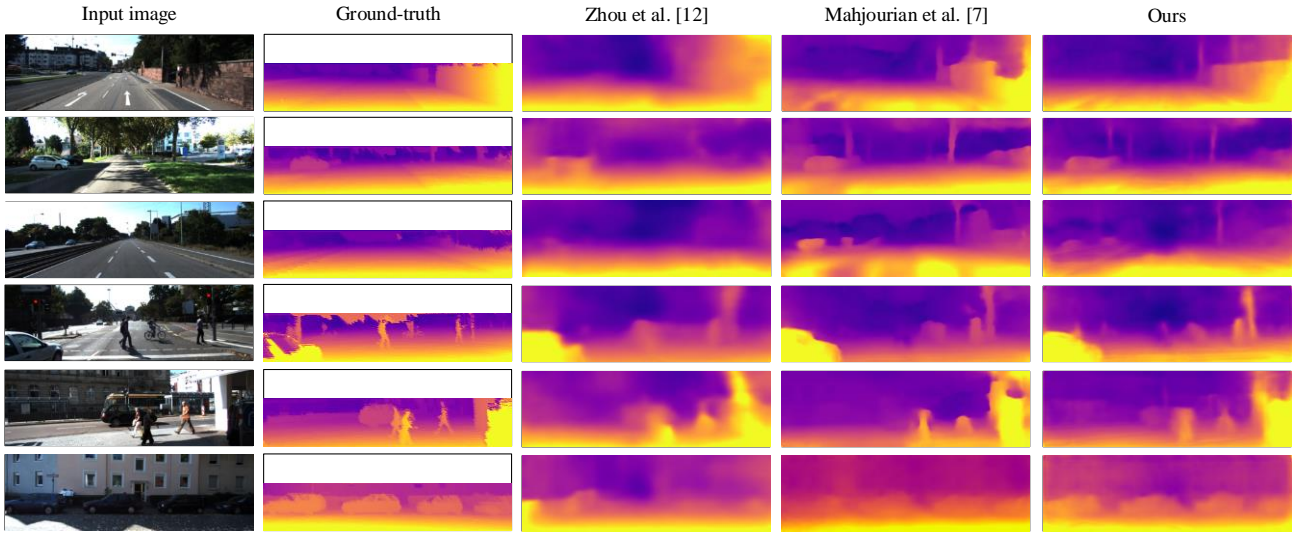

Fig. 8: The comparison of the unsupervised depth estimation between Zhou et al. [12], Mahjourian et al. [7] and ours on the Eigen test set [3]. The ground-truth is interpolated for display. Note that our model is trained without the depth normalization [18] for fair comparison.

TABLE II: Absolute Trajectory Error (ATE) averaged over all multi-frame snippets (lower is better). Ours is calculated over 3-frame snippets. Our model is trained without the depth normalization [18] for fair comparison.

| Method | Seq.09 | Seq.10 |
| --- | --- | --- |
| ORB-SLAM (full) | 0.014 ± 0.008 | 0.012 ± 0.011 |
| ORB-SLAM (short) | 0.064 ± 0.141 | 0.064 ± 0.130 |
| Zhou et al. [12] | 0.021 ± 0.017 | 0.020 ± 0.015 |
| Mahjourian et al. [7] | 0.013 ± 0.010 | 0.012 ± 0.011 |
| GeoNet [16] | 0.012 ± 0.007 | 0.012 ± 0.009 |
| Ours | **0.009 ± 0.005** | **0.008 ± 0.007** |

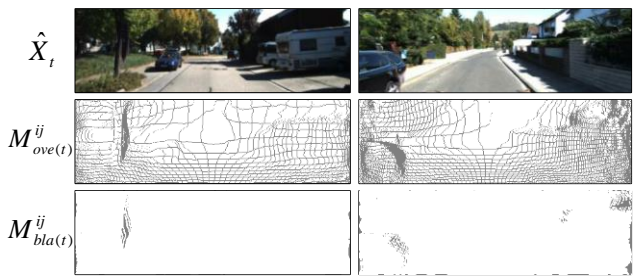

Fig. 9: In the set of images on the left, the extra trees were masked by both masks in the right half that were incorrectly reconstructed due to occlusion. In the set of images on the right, the extra part of the reconstructed car's tail were masked by both masks. However, the overlap mask is denser, and the blank mask is sparse.

Fig. 8 shows a visual comparison of the depth test results. As shown, our results overcome the problem of being too thick for the thin objects such as trees. We think this is mainly because our mask solves the problem of occlusion and the trees are particularly prone to occlusion with the background area due to the thinness. It is shown that ours have fine-grained performance, mainly because the occlusion problem is solved by multiple masks in our method. In addition, ours is more robust to error-prone areas such as sky and roads.

### B. Evaluation of Camera Pose

The ego-motion of the camera is evaluated on the official KITTI odometry split according to the general method [12]. The results for comparation are listed in Table II. The table compares other recent state-of-the-art unsupervised results and ORB-SLAM [24]. ORB-SLAM (full) allows for closed loop and relocation, while ORB-SLAM (short) has no closed loop and relocation. As shown in Table II, our method achieves state-of-the-art in unsupervised monocular method, and exceeds ORB-SLAM (full). We believe that this is because the rich mask information provides guidance for the training of the pose network, so that the pose network learns to ignore the occlusion area for pose estimation.

### C. Visualization of the Overlap Mask and Blank Mask

Although the two masks have similar causes, in practice, the two masks are quite different as shown in Fig. 9. The coverage of the blank mask is smaller than the overlap mask. This is mainly due to the dispersion of pixels and the different mechanism of the two masks.

The overlap mask shows very interesting features. The grid lines of the surface contain the motion information of the camera and the structural information of the space. The fused information may help the learning of the spatial structure or moving objects. We look forward to getting more information from the mask next.

## VI. CONCLUSION

Depth and pose can be estimated only through unsupervised learning of a monocular video stream in our method. We use very natural and simple methods instead of adding edge networks [17] or complex ICP back propagation [7] for training. However, our depth estimation and pose estimation results exceeded them.

In addition, dynamic objects may be the key factors affecting the further improvement of accuracy observed in our experiments, which will be the focus of our future work.